\documentclass[runningheads,a4paper]{llncs}

\usepackage{amssymb}
\usepackage{graphicx}

\usepackage{url}
\urldef{\mailsa}\path| xinsheng.xuan@cripac.ia.ac.cn, { bo.peng, wwang, jdong }@nlpr.ia.ac.cn |

\newcommand{\keywords}[1]{\par\addvspace\baselineskip
\noindent\keywordname\enspace\ignorespaces#1}

\begin{document}

\mainmatter  

\title{On the Generalization of GAN Image Forensics}

\titlerunning{On the Generalization of GAN Image Forensics}

%
%
\author{Xinsheng Xuan, Bo Peng, Wei Wang and Jing Dong\thanks{This work is funded by the National Natural Science Foundation of China (Grant No. 61502496, No. 61303262 and No. U1536120) and Beijing Natural Science Foundation (Grant No. 4164102).}}
\authorrunning{X. Xuan et al.}

\institute{Center for Research on Intelligent Perception and Computing,\\
	National Laboratory of Pattern Recognition, \\
Institute of Automation, Chinese Academy of Sciences, Beijing, China\\
\mailsa\\
}

%
%

\toctitle{Lecture Notes in Computer Science}
\tocauthor{Authors' Instructions}
\maketitle

\begin{abstract}
Recently GAN generated face images are more and more realistic with high-quality, even hard for human eyes to detect. On the other hand, the forensics community keeps on developing methods to detect these generated fake images and try to ensure the credibility of visual contents. Although researchers have developed some methods to detect generated images, few of them explore the important problem of generalization ability of forensics model. As new types of GANs are emerging fast, the generalization ability of forensics models to detect new types of GAN images is absolutely an essential research topic, which is also very challenging. In this paper, we explore this problem and propose to use preprocessed images to train a forensic CNN model. By applying similar image level preprocessing to both real and fake images, unstable low level noise cues are destroyed, and the forensics model is forced to learn more intrinsic features to classify the generated and real face images. Our experimental results also prove the effectiveness of the proposed method.
\end{abstract}

\section{Introduction}

\begin{figure}
	\centering
	\includegraphics[width=0.8\textwidth]{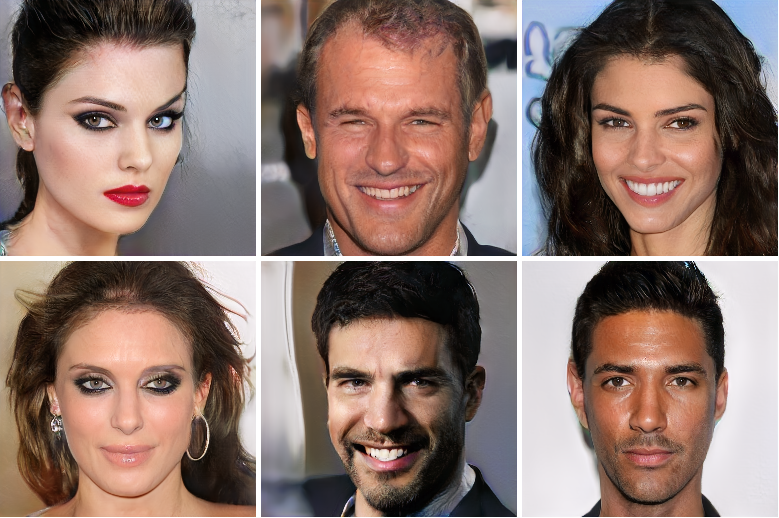}
	\caption{Generated face image examples with PG-GAN \cite{4}.}
	\label{pggan}
\end{figure}

Generative Adversarial Networks (GANs) \cite{goodfellow2014generative} are generative models that learn the distribution of the data without any supervision. Currently, GANs are the most popular and effective generative models for image generation and the generated images could reach very high quality, even human eyes could not tell them apart from real images. Some examples are shown in Fig.1. Owing to the advancement image synthesis of GAN, it also brings a serious forensics problem if we could not distinguish fake image from real ones. For example, DeepFake is a GAN-based technology that can replace a person's face with another person's or animal's face \cite{timmurphy.org}. Criminals can use the generated images to make fake news, and the rumors brought by fake news can have a serious negative impact on our community. In addition, if the generated face can be used to deceive the face recognition system, it will challenge the system security and may cause the collapse of the entire recognition system.

Although there have been some methods proposed in the literature for detecting AI generated images, existing methods are almost exclusively for the detection of one type of generated images, but the detection performance of other unseen types of generated images is not addressed. As new types of GAN models are emerging quickly, the generalization ability of forensics method to other unseen types of generated fake images is becoming more important for the forensic analysis.

The improvement of generalization performance has always been an arduous task. To improve the generalization ability of image forensics model, some primary studies are done in this paper. We adopt a novel method of image preprocessing, e.g. Gaussian Blur and Gaussian Noise, in the training phase to enhance the generalization ability of our forensics Convolutional Neural Network (CNN) model. Our method is quite different from traditional forensics method, the purpose of the general forensic method is to enhance high frequency pixel noise and to focus on the clues in low level pixel statistics. Whereas our work is to destroy or depress these unstable low level high frequency noise cues. The motivation behind using image preprocessing is to improve pixel level statistical similarity between real images and fake images, so that the forensic classifier is forced to learn more intrinsic and meaningful features, rather than the style of the generation model. Hence the classifier will have better generalization ability for the aim of forensic. The experimental results we conduct in this paper also validate the idea of the proposed method.

\section{Related Work}

There are some related work proposed to detect AI generated fake images or videos using deep networks. To detect DeepFake video, different detection methods have been proposed \cite{guera2018deepfake,li2018exposing,li2018ictu,yang2018exposing,3}. In addition, some works focus on the detection of GAN generated images \cite{marra2018detection,tariq2018detecting,1,2}. In \cite{marra2018detection}, the authors present a study on the detection of images translation from GANs. But some of them show dramatic impairments on Twitter-like compressed images. Tariq et al.\cite{tariq2018detecting} use ensemble classifiers to detect fake face images created by GANs. A method based on color statistical features is proposed in \cite{1}, and several detection schemes are designed according to the practicability. Nhu et al.\cite{2} proposed another model based on convolutional neural network to detect gernerated face images, which is based on transfer learning from a deep face recognition network. These image forensics methods can perform well on test dataset that is homologous to the training dataset. 

However, most of the above work do not pay attention to the generalization ability of their forensics models.  They only train and test their methods on the same type of generated image, but the generalization ability to other fake images generated by new GANs models are unknown. An exception is the ForensicTransfer work proposed by Cozzolino et al \cite{cozzolino2018forensictransfer}. The authors use a new autoencoder-based architecture which enforces activations in different parts of a latent vector for the real and fake classes. They devise a learning based forensic detector which adapts well to new domains, and they handle scenarios where only a handful of target domain fake examples are available during training. However, in a real application, we may not have an example images from an unknown generation model. Thus, in this work we propose to improve the generalization ability without using any target domain fake images. 

\section{Proposed Method}

\begin{figure}[htb]
	\centering
	\includegraphics[width=\textwidth]{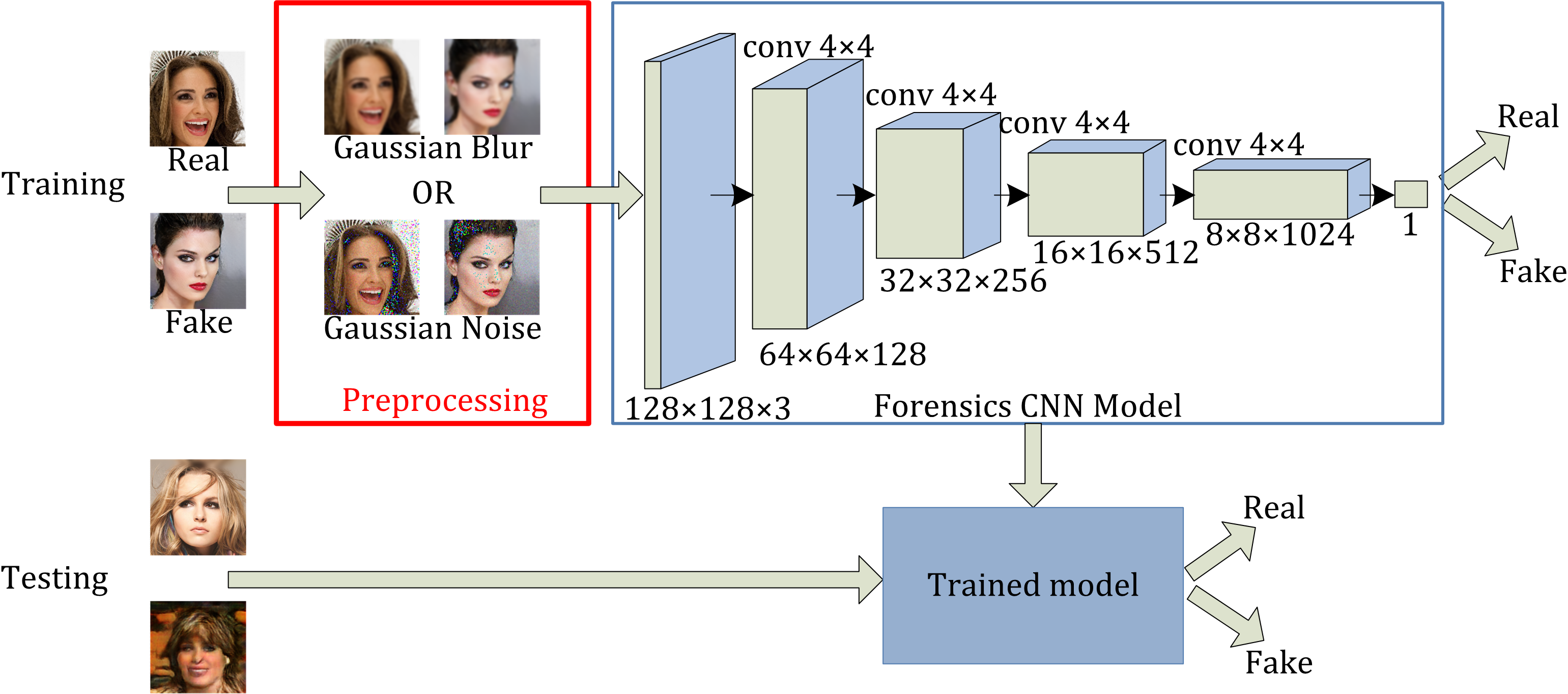}
	\caption{The overall framework of the proposed method.}
	\label{train}
\end{figure}

The improvement of model generalization ability has always been an important and difficult issue, and we propose a method based on image preprocessing to solve this problem. A key difference of our proposed method from other GAN forensics work is that we use an image preprocessing step in the training stage to destroy low level unstable artifact of GAN images and force the forensics discriminator to focus on more intrinsic forensic clues. In this way, our method is a quite different exploration to existing image forensics or image steganalysis networks \cite{qian2015deep,bayar2016deep,yang2016recapture}, where the network is designed to enhance high frequency pixel noise and to focus on the clues in low level pixel statistics. Whereas we intentionally destroy or depress these low level high frequency clues by introducing a preprocessing step using smoothing filtering or noise. By doing this we can improve the low level similarity between real images and fake images, so that the forensic classifier is forced to learn more intrinsic features that have better generalization ability.

From a machine learning perspective, training and testing are two different phases. In the training stage, the training workflow is as shown in Fig.\ref{train}. We add an image preprocessing operation in front of the entire network architecture, where image preprocessing operation can be smoothing filtering or adding noise. In the testing stage, we used the network architecture shown in Fig.\ref{train}. At this stage, we abandon the preprocessing operation, and directly use original images as input. 

In this work, Gaussian blur and Gaussian noise are used as our image preprocessing methods. Adding Gaussian blur and Gaussian noise can both change low level pixel statistics, which serve well for our purpose of depressing low level unstable clues. In order to increase the diversity of training samples, we apply random extent of these preprocessings. The kernel size of Gaussian blur is randomly chosen from 1, 3, 5 and 7 for each training batch. Similarly, the standard deviation of Gaussian noise is randomly set between 0 and 5 for each batch. Note that Gaussian blur of kernel 1 and Gaussian noise of 0 deviation result in no change to the original images.

As our main focus is to verify the effectiveness of proposed preprocessing operation on improving generalization ability, we do not design a complex CNN network architecture. The network architecture of our approach uses a simple DCGAN \cite{6} network's discriminator network. The whole CNN network architecture is shown in Fig.\ref{train}. The input of the network are real and fake images, with image size of 128x128. The network is a binary classifier, with four convolutional layers, and all convolutions have stride 2 and padding 1, and all convolution kernel size is 4x4. For the four convolutional layers, we use the Batch Normalization except the first layer, and use Leaky Rectified Linear Unit activation functions that introduce non-linearities. The loss function and optimization algorithm are Binary Cross Entropy Loss and Adaptive Moment Estimation respectively. 

At test stage, we use the trained CNN model to make forensic decisions on testing images. A difference from the training stage is that we do not preprocess the testing images. This is because the training images also inclue cases of non-preprocessed images from Gaussian blur of kernel 1 and Gaussian noise of deviation 0.

\section{Experiments}

\subsection{Experimental Setups}

For the real face image dataset, we use the CelebA-HQ \cite{4}, which contains high quality face images of 1024x1024 resolution. We denote the real images in CelebA-HQ as $R_{cel}$. As for fake datasets, we use images generated by DCGAN \cite{6}, WGAN-GP \cite{7} and PGGAN \cite{4}, and they are respectively denoted as $F_{dc}$, $F_{wg}$ and $F_{pg}$. For DCGAN and WGAN-GP, we first train the generative models using CelebA \cite{liu2015deep} dataset, and then use these trained GAN models to generate fake face images. The PGGAN model is a high quality image generation model based on progressive growing. Due to the long training time of the PGGAN model, we directly download fake image dataset provided by authors \cite{4}. The size of images generated by DCGAN and WGAN-GP models is 128x128, and this is the input image size that our CNN model requires. However, the size of both real images and PG-GAN generated images is of high resolution 1024x1024, so we resize them to 128x128. 

In our experiments, we train our CNN forensics model on only $F_{pg}$ and $R_{cel}$ datasets, and the rest two generated datasets $F_{dc}$ and $F_{wg}$ are just used for testing the generalization ability of trained model. Here images in $F_{dc}$ and $F_{wg}$ are treated as unseen generated images from new GANs that are different from the training data. The $F_{pg}$ and $R_{cel}$ datasets each has 20K images, where the first 10K images are used for model training and the last 10K for testing. The $F_{dc}$ and $F_{wg}$ datasets each contains 10K images for testing generalization ability.

The model trained on $R_{cel}$ and $F_{pg}$ without any image preprocessing is denoted as $M$. For the other two models, the training dataset is processed by Gaussian blur or Gaussian noise, which respectively are denoted as $M_{GB}$ and $M_{GN}$. Then, we use testing images in $R_{cel}$, $F_{pg}$, $F_{dc}$ and $F_{wg}$ to test $M$, $M_{GB}$ and $M_{GN}$ separately. The performance is measured by overall accuracy (ACC), true positive rate (TPR) and true negative rate (TNR), where positive means real images and negative means fake images. 

\subsection{Improvement of Model Generalization}

\begin{table}
	\begin{center}
		\caption{Detection results of models in different preprocessing operations.}\label{preprocessing results}
		\begin{tabular}{|l|l|l|l|l|l|}
			\hline
			No. &  Detector model & Testing set& ACC(\%)& TPR(\%)& TNR(\%)\\
			
			$1$& $M$& $F_{pg}$ + $R_{cel}$& 95.45& 95.12& 95.77\\
			$2$& $M_{GB}$& $F_{pg}$ + $R_{cel}$& 94.28& 93.08& 95.47\\
			$3$& $M_{GN}$& $F_{pg}$ + $R_{cel}$& 95.02& 94.65& 95.38\\ \hline
			
			$4$& $M$& $F_{wg}$ + $R_{cel}$& 64.62& 95.12& 34.12\\
			$5$& $M_{GB}$& $F_{wg}$ + $R_{cel}$& 68.07& 93.08& \bfseries{43.06}\\
			$6$& $M_{GN}$& $F_{wg}$ + $R_{cel}$& 68.28& 94.65& \bfseries{41.91}\\ \hline
			
			$7$& $M$& $F_{dc}$ + $R_{cel}$& 60.55& 95.12& 25.98\\
			$8$& $M_{GB}$& $F_{dc}$ + $R_{cel}$& 64.05& 93.08& \bfseries{35.02}\\
			$9$& $M_{GN}$& $F_{dc}$ + $R_{cel}$& 66.38& 94.65& \bfseries{38.11}\\
			\hline
			
		\end{tabular}
	\end{center}
\end{table}

The experimental results are shown in Table.\ref{preprocessing results}. The experiment is divided into three parts by different test datasets. The test datasets of the first row to the third row are $F_{pg}$ and $R_{cel}$, the test datasets of the fourth row to the sixth row are $F_{wg}$ and $R_{cel}$, and the test datasets of the seventh row to the ninth row are $F_{dc}$ and $R_{cel}$. Compare the $M$ model without image preprocessing operation on row 1 and the $M_{GB}$ and $M_{GN}$ models with image preprocessing operations on row 2 and row 3, ACC, TPR and TNR are almost constant after adding preprocessing. And this means that the proposed method does not damage the model and is relatively stable. From the first three row, it can also be seen that testing on the data which is from the same domain as training data can achieve very high classification performance.

From the data of row 1 we can observe that the detection ACC, TPR and TNR are all higher than 95\% on testing dataset of the same type as training dataset, but the ACC and TNR on rows 4 and 7 are both significantly lower than those in row 1. This result means that the generalization ability of the model on unseen types of fake images is bad. 

Compare row 4 and rows 5, 6 in Table.\ref{preprocessing results}, with the test dataset is $F_{wg}$ and $R_{cel}$, we can see that our trained model can improve the TNR by around 10 percents and the overall ACC is also improved. And this can show that the method of preprocessing operation is effective for improving generalization ability on unseen generated images. Similarly, comparing row 7 and rows 8, 9, TNR also has an improvement of about 10 percents. Although the performance increment is not all that large due to the inherent difficulty of this problem, it is sufficient to show that our methods can improve generalization ability on unseen types of fake image datasets. 

\begin{figure}[htb]
	\centering
	\includegraphics[width=0.7\textwidth]{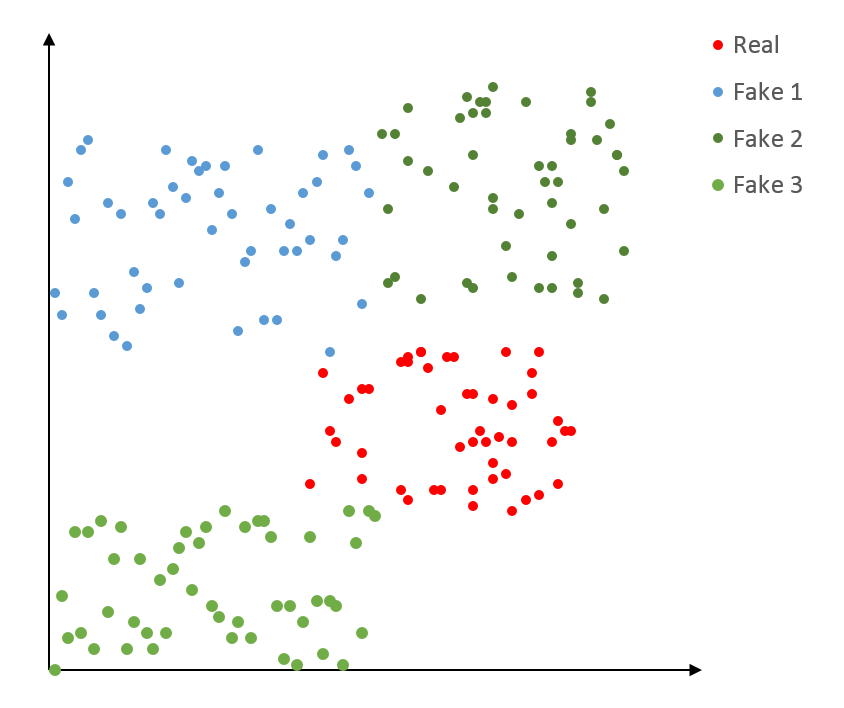}
	\caption{Possible distribution of real and fake images simplified on a two-dimensional feature space. }
	\label{summary}
\end{figure}

Different from other forensic methods, our method uses an image preprocessing to suppress unstable noise cues. From the experimental results, it can be seen that the proposed image preprocessing method can actually lead to a certain improvement of generalization, although the increment is not quite large. After our analysis, we believe that the reason for the difficulty in forensics generalization may be as shown in Fig.\ref{summary}. There are many types of generator models, and likely to be more in the future, and the distribution of images generated by each model may vary greatly. As shown in Fig.\ref{summary}, Fake1, Fake2 and Fake3 belong to differently distributed fake images. Although they are all fake images, the distribution difference between them is quite large. Therefore, to train a forensics model which can generalize to future unknown generated fake images is a very challenging task. We hope researchers can carry on in this line of research in the future to develop more and more effective solutions. 

\section{Conclusion}

In this paper, we have investigated the issue of generalization ability of detection model for generated images. Perhaps because of the difficulty of generalization capabilities improvement, we found that most of the existing detection models did not pay attention to the improvement of generalization capabilities. Based on the observations, we propose to improve the generalization ability of a CNN forensics model by adding an image preprocessing step before training to force the discriminator to learn more intrinsic and generalizable features. To evaluate the performance of the proposed method, extensive experiments have been conducted. The experimental results show that our approach is effective in improving generalization, although the performance increment is not all that large due to the inherent difficulty of this problem. Observed from the experiments, the distribution of fake images generated by different models may be quite different. In short, the improvement of generalization is quite difficult, and we take a very different strategy compared to existing, but only achieved some preliminary results. We hope to inspire more work in this direction. In the future, we will continue to optimize the generalization of the detection model in other ways.

\bibliographystyle{splncs}
\bibliography{Ref}{}

\end{document}